\documentclass[runningheads]{llncs}
\pdfoutput=1
\usepackage{amsmath}
\usepackage{booktabs} 
\usepackage{caption} 
\usepackage{subcaption} 
\usepackage{graphicx}
\usepackage{pgfplots}
\usepackage[all]{nowidow}
\usepackage[utf8]{inputenc}
\usepackage[round]{natbib}
\usepackage{tikz}
\usetikzlibrary{er,positioning,bayesnet}
\usepackage{multicol}
\usepackage{algpseudocode,algorithm,algorithmicx}
\usepackage{hyperref}
\usepackage[inline]{enumitem} 
\usepackage[bottom]{footmisc}
\definecolor{blue}{HTML}{1F77B4}
\definecolor{orange}{HTML}{FF7F0E}
\definecolor{green}{HTML}{2CA02C}

\pgfplotsset{compat=1.14}

\begin{document}
\title{Diachronic Topics in New High German Poetry}
%
%
\author{Thomas N. Haider\inst{1,2}}
%
%
\institute{Department of Language and Literature\\
  Max Planck Institute for Empirical Aesthetics, Frankfurt\\
   \and
Institut für Maschinelle Sprachverarbeitung (IMS) \\ University of Stuttgart \\
\email{\texttt{thomas.haider@ae.mpg.de}}}
\maketitle              
%
%



\begin{abstract}
Statistical topic models are increasingly and popularly used by Digital Humanities scholars to perform distant reading tasks on literary data \citep{navarro2018poetic}, \citep{hettinger2016classification}. It allows us to estimate what people talk about. Especially Latent Dirichlet Allocation (LDA), see \citep{blei2003latent}, has shown its usefulness, as it is unsupervised, robust, easy to use, scalable, and it offers interpretable results. In a preliminary study, we apply LDA to a corpus of New High German poetry (textgrid, with 51k poems, 8m token) and interpret salient topics, their trend over time (1575–1925 A.D.), and use the distribution of topics over documents for a classification of poems into time periods and for authorship attribution.\footnotetext[0]{Proceedings of the International Digital Humanities Conference DH2019, Utrecht \\ \url{https://dev.clariah.nl/files/dh2019/boa/1031.html}}
\end{abstract}

\section{Corpus}
The Digital Library in the TextGrid Repository represents an extensive collection of German texts in digital form \citep{vanscheidttextgrid}. It was mined from \url{http://zeno.org} and covers a time period from the mid 16th century up to the first decades of the 20th century. It contains many important texts that can be considered as part of the literary canon, even though it is far from complete (e.g. it contains only half of Rilke’s work). We find that around 51k texts are annotated with the label ’verse’ (TGRID-V), not distinguishing between ’lyric verse’ and ’epic verse’. However, the average length of these texts is around 150 token, dismissing most epic verse tales. Also, the poems are distributed over 229 authors, where the average author contributed 240 poems (median 131 poems). A drawback of TGRID-V is the circumstance that it contains a noticeable amount of French, Dutch and Latin (over 400 texts). To constrain our dataset to German, we filter foreign language material with a stopword list, as training a dedicated language identification classifier is far beyond the scope of this work.

\begin{figure}
    \centering
    \includegraphics[width=\textwidth]{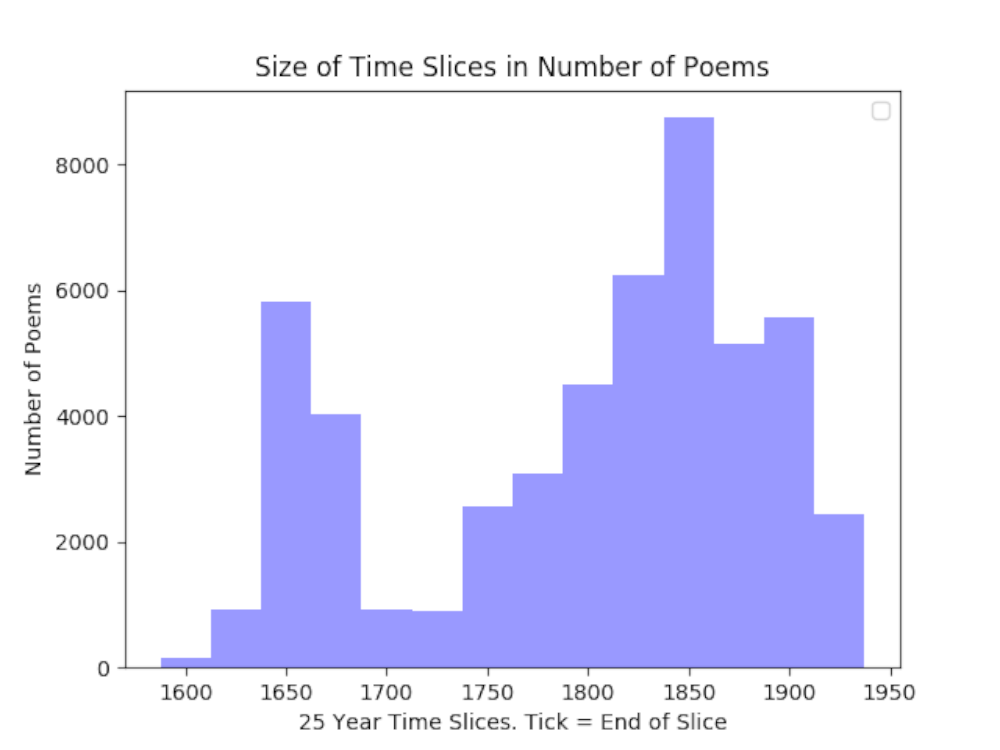}
    \caption{25 year Time Slices of Textgrid Poetry (1575--1925)}
    \label{fig:my_label}
\end{figure}

\section{Experiments}
We approach diachronic variation of poetry from two perspectives. First, as distant reading task to visualize the development of clearly interpretable topics over time. Second, as a downstream task, i.e. supervised machine learning task to determine the year (the time-slot) of publication for a given poem. We infer topic distributions over documents as features and pit them against a simple style baseline.

We use the implementation of LDA as it is provided in genism \citep{vrehuuvrek2011gensim}. LDA assumes that a particular document contains a mixture of few salient topics, where words are semantically related. We transform our documents (of wordforms) to a bag of words representation, filter stopwords (function words), and set the desired number of topics=100 and train for 50 epochs to attain a reasonable distinctness of topics. We choose 100 topics (rather than a lower number that might be more straightforward to interpret) as we want to later use these topics as features for downstream tasks. We find that wordforms (instead of lemma) are more useful for poetry topic models, as these capture style features (rhyme), orthographic variations ('hertz' instead of 'herz'), and generally offer more interpretable results.

\subsection{Topic Trends}
We retrieve the most important (likely) words for all 100 topics and interpret these (sorted) word lists as aggregated topics, e.g. topic 27 (figure 2) contains: Tugend (virtue), Kunst (art), Ruhm (fame), Geist (spirit), Verstand (mind) and Lob (praise). This topic as a whole describes the concept of ’artistic virtue’.

In certain clusters (topics) we find poetic residuals, such that rhyme words often cluster together (as they stand in proximity), e.g. topic 52 with: Mund (mouth), Grund (cause, ground), rund (round).

To discover trends of topics over time, we bin our documents into time slots of 25 years width each. See figure 1 for a plot of the number of documents per bin. The chosen binning slots offer enough documents per slot for our experiments. To visualize trends of singular topics over time, we aggregate all documents d in slot s and add the probabilities of topic t given d and divide by the number of all d in s. This gives us the average probability of a topic per timeslot. We then plot the trajectories for each single topic. See figures 2–6 for a selection of interpretable topic trends. Please note that the scaling on the y-axis differ for each topic, as some topics are more pronounced in the whole dataset overall.

\begin{figure}
    \centering
    \includegraphics[width=\textwidth]{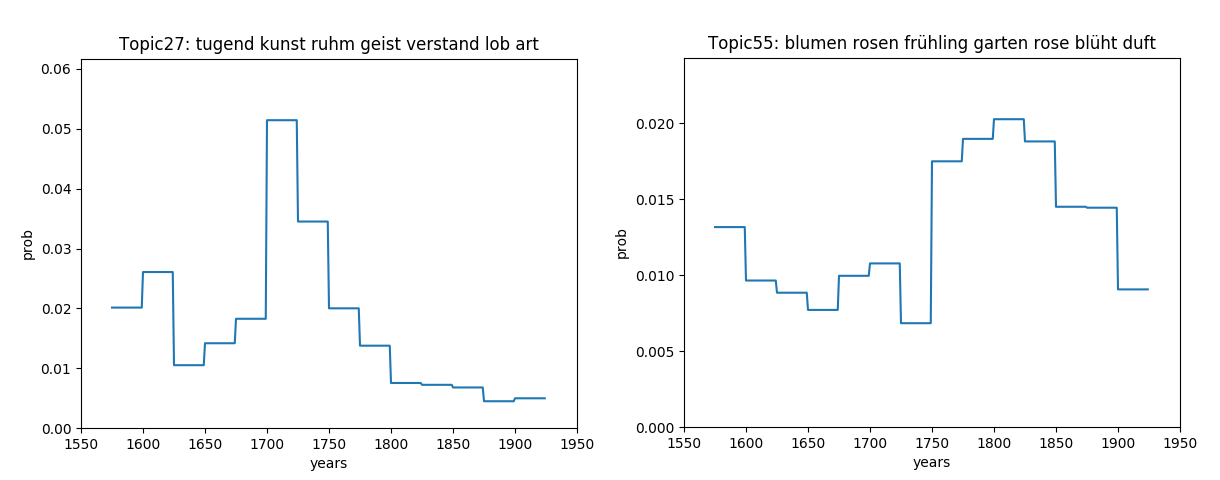}
    \caption{left: Topic 27 'Virtue, Arts' (Period: Enlightenment), right: Topic 55 'Flowers, Spring, Garden' (Period: Early Romanticism)}
    \label{fig:my_label}
\end{figure}

\begin{figure}
    \centering
    \includegraphics[width=\textwidth]{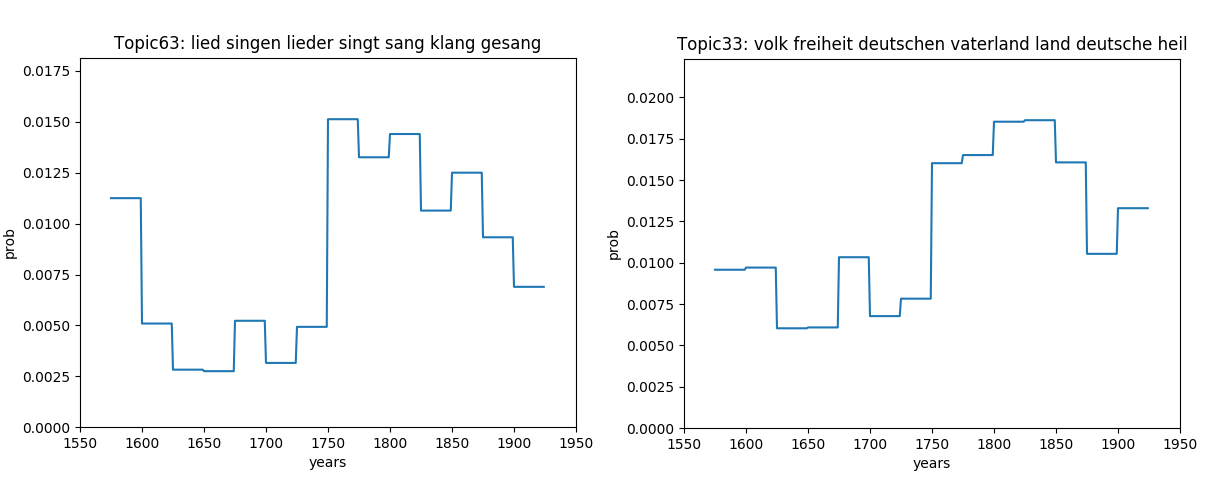}
    \caption{left: Topic 63 'Song' (Period: Romanticism), right: Topic 33 'German Nation' (Period: Vormärz, Young Germany))}
    \label{fig:my_label}
\end{figure}

\begin{figure}
    \centering
    \includegraphics[width=\textwidth]{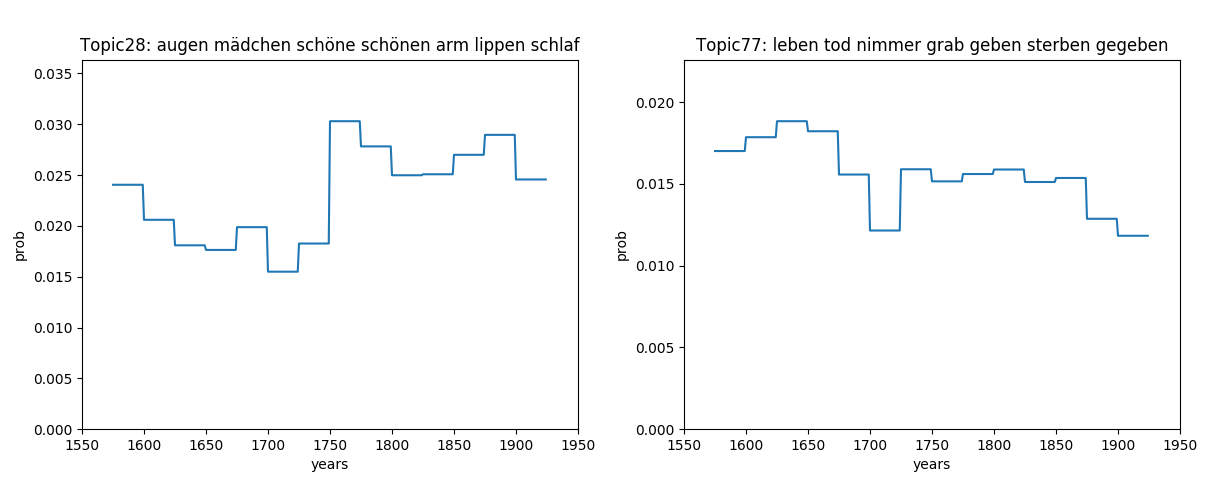}
    \caption{left: Topic 28 'Beautiful Girls' (Period: Omnipresent, Romanticism), right: Topic 77 'Life \& Death' (Period: Omnipresent, Barock}
    \label{fig:my_label}
\end{figure}

\begin{figure}
    \centering
    \includegraphics[width=\textwidth]{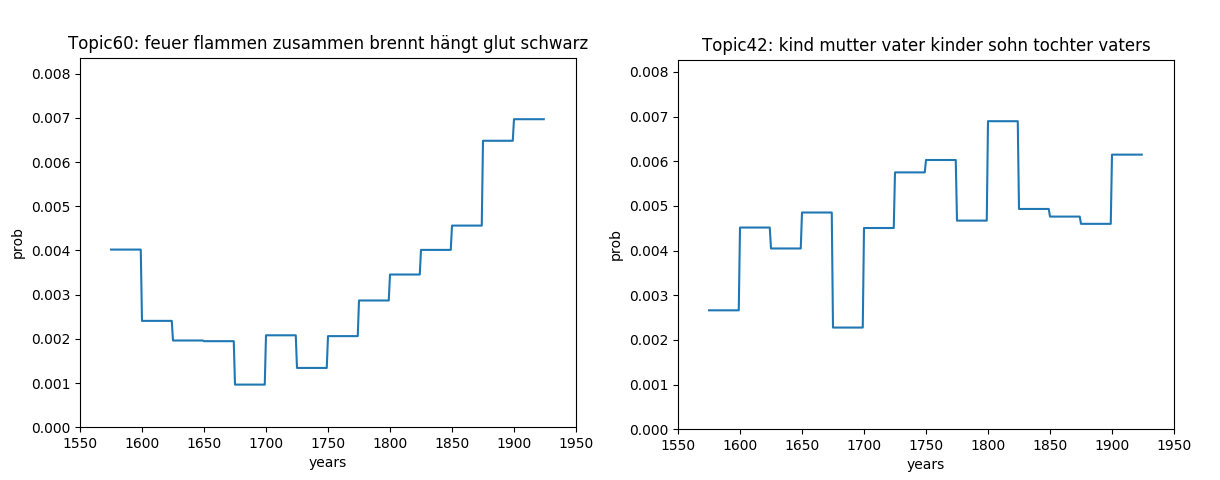}
    \caption{ left: Topic 60 'Fire' (Period: Modernity), right: Topic 42 'Family' (no period, fluctuating)}
    \label{fig:my_label}
\end{figure}

\begin{figure}
    \centering
    \includegraphics[width=\textwidth]{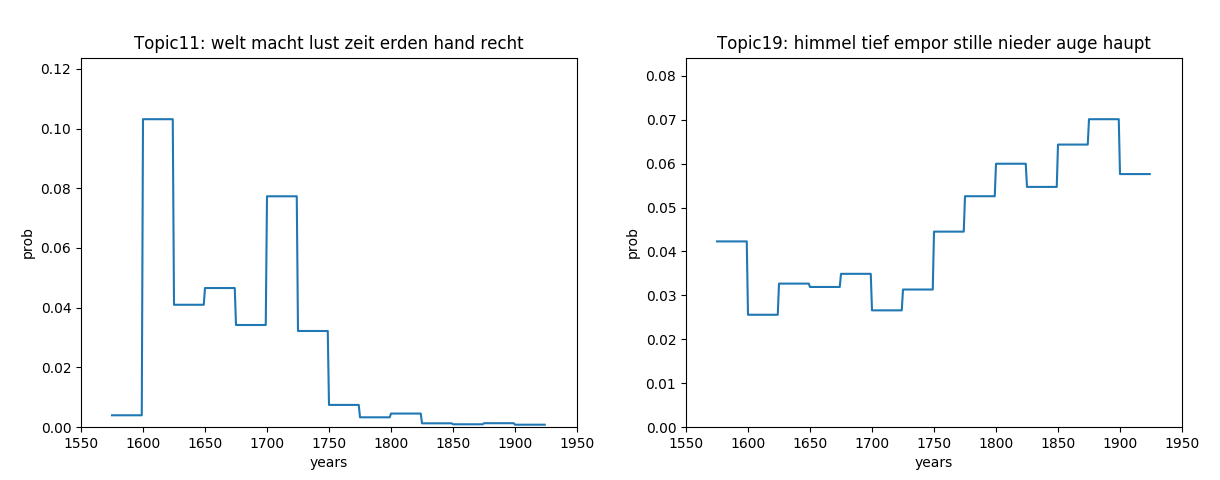}
    \caption{ Most informative topics for classification; left: Topic 11 'World, Power, Lust, Time' (Period: Barock), right: Topic 19 'Heaven, Depth, Silence' (Period: Romanticism, Modernity)}
    \label{fig:my_label}
\end{figure}

Some topic plots are already very revealing. The topic ‘artistic virtue’ (figure 2, left) shows a sharp peak around 1700—1750, outlining the period of Enlightenment.
Several topics indicate Romanticism, such as ‘flowers’ (figure 2, right), ‘song’ (figure 3, left) or ‘dust, ghosts, depths’ (not shown). The period of 'Vormärz' or 'Young Germany' is quite clear with the topic ‘German Nation’ (figure 3, right). It is however hardly distinguishable from romantic topics.

We find that the topics 'Beautiful Girls' (figure 4, left) and 'Life \& Death' (figure 4, right) are always quite present over time, while 'Girls' is more prounounced in Romanticism, and 'Death' in Barock.

We find that the topic 'Fire' (figure 5, left) is a fairly modern concept, that steadily rises into modernity, possibly because of the trope 'love is fire'. Next to it, the topic 'Family' (figure 5, right) shows wild fluctuation over time.

Finally, figure 6 shows topics that are most informative for the downstream classification task: Topic 11 'World, Power, Time' (left) is very clearly a Barock topic, ending at 1750, while topic 19 'Heaven, Depth, Silence' is a topic that rises from Romanticism into Modernity.

\subsection{Classification of Time Periods and Authorship}
To test whether topic models can be used for dating poetry or attributing authorship, we perform supervised classification experiments with Random Forest Ensemble classifiers. We find that we obtain better results by training and testing on stanzas instead of full poems, as we have more data available. Also, we use 50 year slots (instead of 25) to ease the task.

For each document we determine a class label for a time slot. The slot 1575–1624 receives the label 0, the slot 1625–1674 the label 1, etc.. In total, we have 7 classes (time slots).

As a baseline, we implement rather straightforward style features, such as line length, poem length (in token, syllables, lines), cadence (number of syllables of last word in line), soundscape (ratio of closed to open syllables, see \citep{hench2017phonological}), and a proxy for metre, the number of syllables of the first word in the line.

We split the data randomly 70:30 training:testing, where a 50:50 shows (5 points) worse performance. We then train Random Forest Ensemble classifiers and perform a grid search over their parameters to determine the best classifier. Please note that our class sizes are quite imbalanced.

The Style baseline achieves an Accuracy of 83\%, LDA features 89\% and a combination of the two gets 90\%. However, training on full poems reduces this to 42—52\%.

The most informative features (by information gain) are: Topic11 (.067), Topic 37 (.055), Syllables Per Line (.046), Length of poem in syllables (.031), Topic19 (.029), Topic98 (.025), Topic27 ('virtue') (.023), and Soundscape (.023).

For authorship attribution, we also use a 70:30 random train:test split and use the author name as class label. We only choose the most frequent 180 authors. We find that training on stanzas gives us 71\% Accuracy, but when trained on full poems, we only get 13\% Accuracy. It should be further investigated is this is only because of a surplus of data.

\subsection{Conclusion \& Future Work}
We have shown the viability of Latent Dirichlet Allocation for a visualization of topic trends (the evolution of what people talk about in poetry). While most topics are easily interpretable and show a clear trend, others are quite noisy. For an exploratory experiment, the classification into time slots and for authors attribution is very promising, however far from perfect. It should be investigated whether using stanzas instead of whole poems only improves results because of more available data. Also, it needs to be determined if better topic models can deliver a better baseline for diachronic change in poetry, and if better style features will outperform semantics. Finally, only selecting clear trending and peaking topics (through co-variance) might further improve the results.
%
%
%
\bibliographystyle{unsrtnat}
\bibliography{biblio}
\end{document}